%% file: main.tex
\author{\IEEEauthorblockN{1\textsuperscript{st} Given Name Surname}
\IEEEauthorblockA{\textit{dept. name of organization (of Aff.)} \\
\textit{name of organization (of Aff.)}\\
City, Country \\
email address or ORCID}
\and
\IEEEauthorblockN{2\textsuperscript{nd} Given Name Surname}
\IEEEauthorblockA{\textit{dept. name of organization (of Aff.)} \\
\textit{name of organization (of Aff.)}\\
City, Country \\
email address or ORCID}
}

\documentclass[conference]{IEEEtran}
\IEEEoverridecommandlockouts

\usepackage{cite}
\usepackage{amsmath,amssymb,amsfonts}
\usepackage{graphicx}
\usepackage{textcomp}
\usepackage{xcolor}
\usepackage{algorithm}
\usepackage{algpseudocode}
\usepackage{booktabs}
\usepackage{hyperref}
\newcommand{\boldheader}[1]{\noindent\textbf{#1}}

\def\BibTeX{{\rm B\kern-.05em{\sc i\kern-.025em b}\kern-.08em
    T\kern-.1667em\lower.7ex\hbox{E}\kern-.125emX}}
    
\begin{document}

\title{ACT360: An Efficient 360-Degree Action Detection and Summarization Framework for Mission-Critical Training and Debriefing}


\author{\IEEEauthorblockN{Aditi Tiwari, Klara Nahrstedt}
\IEEEauthorblockA{University of Illinois Urbana-Champaign (UIUC)\\
\{aditit5, klara\}@illinois.edu}
}

\maketitle

\input{sections/0-abstract}
\input{sections/1-introduction}

\input{sections/2-backgroundrelatedwork}

\input{sections/3-framework}
\input{sections/4-dataset}

\input{sections/5-experimentresults}

\section{Conclusion} \label{sec:conclusion}

ACT360 presents a novel framework for mission-critical training and debriefing using 360-degree video and machine learning, addressing challenges such as panoramic distortions through Equirectangular-Aware Convolution (EAC) and 360-degree spatial attention. The core action detection model, 360YOWO, balances accuracy and efficiency, achieving 0.850 mAP while reducing model size by 74\% and improving inference speed by 31.25\% through quantization and pruning, enabling deployment on resource-constrained hardware. Integrated with 360AIE, ACT360 enhances post-training debriefing through automated action retrieval and textual summarization. While ACT360 significantly improves over existing 2D models adapted for 360-degree video, the lack of dedicated 360-degree action detection benchmarks remains a challenge. Future work could explore self-supervised learning, multi-camera fusion, and robustness to low-visibility conditions to enhance generalizability across mission-critical applications.

\section{Acknowledgments}
We are deeply grateful for the support from the firefighters and researchers at the Illinois Fire Service Institute (IFSI), especially Dr. Farzaneh Masoud and Dr. Dac Nguyen, who helped us understand firefighter training and assisted in the collection of the dataset. I would also like to extend my sincere thanks to Dr. Zhisheng Yan for his insightful inputs that enhanced my understanding of 360-degree videos. This research is supported by the National Science Foundation under grant number NSF IIS 21-40645.



\bibliographystyle{IEEEtran}  
\bibliography{references}  

\end{document}

%% file: sections/0-abstract.tex
\begin{abstract} Effective training and debriefing are critical in high-stakes, mission-critical environments such as disaster response, military simulations, and industrial safety, where precision and minimizing errors are paramount. The traditional post-training analysis relies on manually reviewing 2-dimensional (2D) videos, a time-consuming process that lacks comprehensive situational awareness. To address these limitations, we introduce ACT360, a system that leverages 360-degree videos and machine learning for automated action detection and structured debriefing. ACT360 integrates \textit{360YOWO}, an enhanced You Only Watch Once (YOWO) model with spatial attention and equirectangular-aware convolution (EAC) to mitigate panoramic video distortions.

To enable deployment in resource-constrained environments, we apply quantization and model pruning, reducing the model size by 74\% while maintaining robust accuracy (mAP drop of only 1.5\%, from 0.865 to 0.850) and improving inference speed. We validate our approach on a publicly available dataset of 55 labeled 360-degree videos covering seven key operational actions, recorded across various real-world training sessions and environmental conditions. Additionally, ACT360 integrates \textit{360AIE (Action Insight Explorer)}, a web-based interface for automatic action detection, retrieval, and textual summarization using large language models (LLMs), significantly enhancing post-incident analysis efficiency.

ACT360 serves as a generalized framework for mission-critical debriefing, incorporating EAC, spatial attention, summarization, and model optimization. These innovations apply to any training environment requiring lightweight action detection and structured post-exercise analysis. \end{abstract}

\begin{IEEEkeywords}
360-degree video analysis, mission-critical training, action detection, machine learning, spatial attention, non-maximum suppression, confidence thresholding, quantization, pruning, latency-aware system, data labeling, large language models, summarization
\end{IEEEkeywords}

%% file: sections/1-introduction.tex
\section{Introduction}

Mission-critical training, including firefighter drills, military exercises, and emergency response simulations, relies on accurate debriefing to evaluate performance and enhance decision-making. Traditionally, training sessions are recorded using fixed 2D cameras, requiring instructors to manually review hours of footage to identify key actions and mistakes. This manual review process is time-consuming, error-prone, and lacks comprehensive situational awareness. In high-stakes environments where every second matters, there is a pressing need for automated action detection and summarization to improve training efficiency and effectiveness.

360-degree videos provide a promising alternative by capturing the entire operational environment, eliminating blind spots, and enabling immersive analysis. However, analyzing 360-degree videos presents significant challenges that existing 2D action detection methods fail to address. Fine-grained action recognition is required to detect complex interactions rather than just identifying objects or individuals. Geometric distortions from equirectangular projection (ERP) introduce stretching and warping, particularly near the poles, which degrade the accuracy of traditional convolutional networks. Additionally, the higher resolution of 360-degree videos significantly increases computational overhead, making efficient processing difficult. Despite advances in deep learning-based action detection models like You Only Watch Once (YOWO) \cite{rwPara1ref5Kpkl2019YouOW}, standard approaches do not account for ERP distortions, leading to inaccurate action localization. Recent works such as InternVideo2 \cite{internvideo} and 2D Conv-RBM + LSTM \cite{convRBM} have improved temporal action localization and spatiotemporal learning but are designed for rectilinear videos and require extensive adaptation for panoramic data. Similarly, TokenLearner \cite{tokenlearner} improves computational efficiency in video classification but lacks the spatial localization capabilities necessary for detecting fine-grained actions in dynamic environments. These limitations highlight the need for a specialized framework capable of addressing the unique challenges of 360-degree action detection.

Furthermore, mission-critical environments require a lightweight, resource-efficient solution that can run on standard computing hardware without relying on large-scale cloud processing. Many existing models prioritize computational efficiency for classification (TokenLearner), hand-object interaction (EffHandNet) \cite{effhandnet}, or general video understanding (InternVideo2), but none are designed for resource-efficient, end-to-end 360-degree action detection. \textit{Therefore, adapting existing 2D models or classification-based frameworks is not a viable solution for accurate action detection in resource-constrained 360-degree mission-critical training scenarios.}

To address these limitations, we present ACT360, a novel debriefing and action detection system for mission-critical training. ACT360 integrates multiple components: (1) 360YOWO (Enhanced YOWO for 360-degree video), a custom action detection model incorporating Equirectangular-Aware Convolution (EAC) and 360-degree Spatial Attention to mitigate ERP distortions and improve accuracy; (2) 360AIE (Action Insight Explorer), a web-based debriefing interface that provides automatic action detection, retrieval, and textual summarization of key events in training videos using large language models (LLMs); (3) Optimized Model Deployment, where quantization and pruning significantly reduce computational complexity, enabling deployment on resource-constrained hardware such as standard laptops with consumer GPUs; (4) We provide a publicly available 360-degree firefighter training dataset with 55 annotated videos of real-world firefighter drills, including ladder operations, door breaching, and hose deployment, available on the project website.

While 360YOWO is validated on firefighter training, the techniques developed in this system can be extended to various fields, including disaster response, military simulations, medical training, and law enforcement. By curating and annotating new datasets, ACT360 can be adapted to other mission-critical training environments, ensuring that action detection remains accurate, efficient, and deployable on resource-constrained devices.

\boldheader{Key Innovations.} This work introduces the following key innovations:
\begin{enumerate}
    \item \textbf{Custom 360-Degree Firefighting Dataset:} We introduce a novel dataset of annotated 360-degree firefighting videos, capturing diverse actions across various environmental conditions. Developed with firefighting instructors, it provides a foundation for action detection in emergency response. Unlike existing datasets such as 360Action \cite{360actiondataset}, which focus on general actions, ours specifically targets mission-critical firefighter training.    
    \item \textbf{Enhanced YOWO for 360-Degree Video (360YOWO):} We introduce 360YOWO, a custom action detection model incorporating EAC and 360-degree Spatial Attention to handle the unique challenges of equirectangular projection (ERP). This enables the model to focus on relevant regions within the panoramic view while mitigating distortions inherent in 360-degree footage. While attention mechanisms have been used in viewport prediction, 360YOWO is specifically designed for fine-grained action detection, identifying subtle yet critical actions in dynamic, high-risk scenarios.
    \item \textbf{Equirectangular-Aware Convolution (EAC):} We develop a custom convolution operation that dynamically adjusts based on latitude coordinates, effectively addressing the warping effects near the poles of equirectangular projections and enabling more accurate feature extraction across the entire 360-degree field of view. EAC significantly improves detection accuracy compared to standard convolutions.
    \item \textbf{Action Insight Explorer (360AIE) with LLM Summarization:} We introduce 360AIE, a debriefing interface that summarizes detected actions using large language models (LLMs). This represents the first integration of LLM-based summarization for 360-degree video in mission-critical settings.
    \item \textbf{Optimized Model Deployment:} We optimize 360YOWO through quantization and pruning, reducing model size while maintaining accuracy. The optimized model can run on a single consumer-grade GPU, making it accessible to a wider range of users, including firefighter training centers and first responder facilities.
\end{enumerate}

%% file: sections/2-backgroundrelatedwork.tex
\section{Background - Definitions, and Related Work}
\label{sec:background_motivation}

\subsection{Definitions}

\boldheader{Action.} An action refers to an observable activity performed by an individual or group in a mission-critical training scenario. Actions range from basic movements (e.g., kneeling) to complex sequences (e.g., deploying a fire hose). The detection granularity depends on the training context and the debriefing requirements.

\boldheader{Mission-Critical.} A mission-critical scenario involves high-risk operations where delayed or incorrect actions can lead to severe consequences. These include firefighting, emergency response, military operations, and industrial safety, requiring rapid coordination under constrained environments.

\boldheader{360-Degree Video: Opportunities and Challenges.} 360-degree video enhances mission-critical training by providing comprehensive situational awareness, eliminating blind spots, and enabling a complete review of trainee actions. It improves training analysis by capturing all interactions for detailed performance evaluations and offers an immersive learning experience, allowing trainees to observe scenarios from multiple perspectives. However, processing 360-degree video presents challenges. Equirectangular Projection (ERP) distortions introduce spatial warping, particularly near the poles, affecting convolutional neural network (CNN)-based models \cite{eac}. The computational complexity is significantly higher than standard 2D videos, given the fourfold increase in resolution, making efficient processing a challenge. Moreover, there is a lack of specialized models for 360-degree action detection, as most existing methods are designed for rectilinear video and fail to account for panoramic distortions and altered spatial relationships.

\begin{figure}[t]
    \centering
    \includegraphics[width=\linewidth]{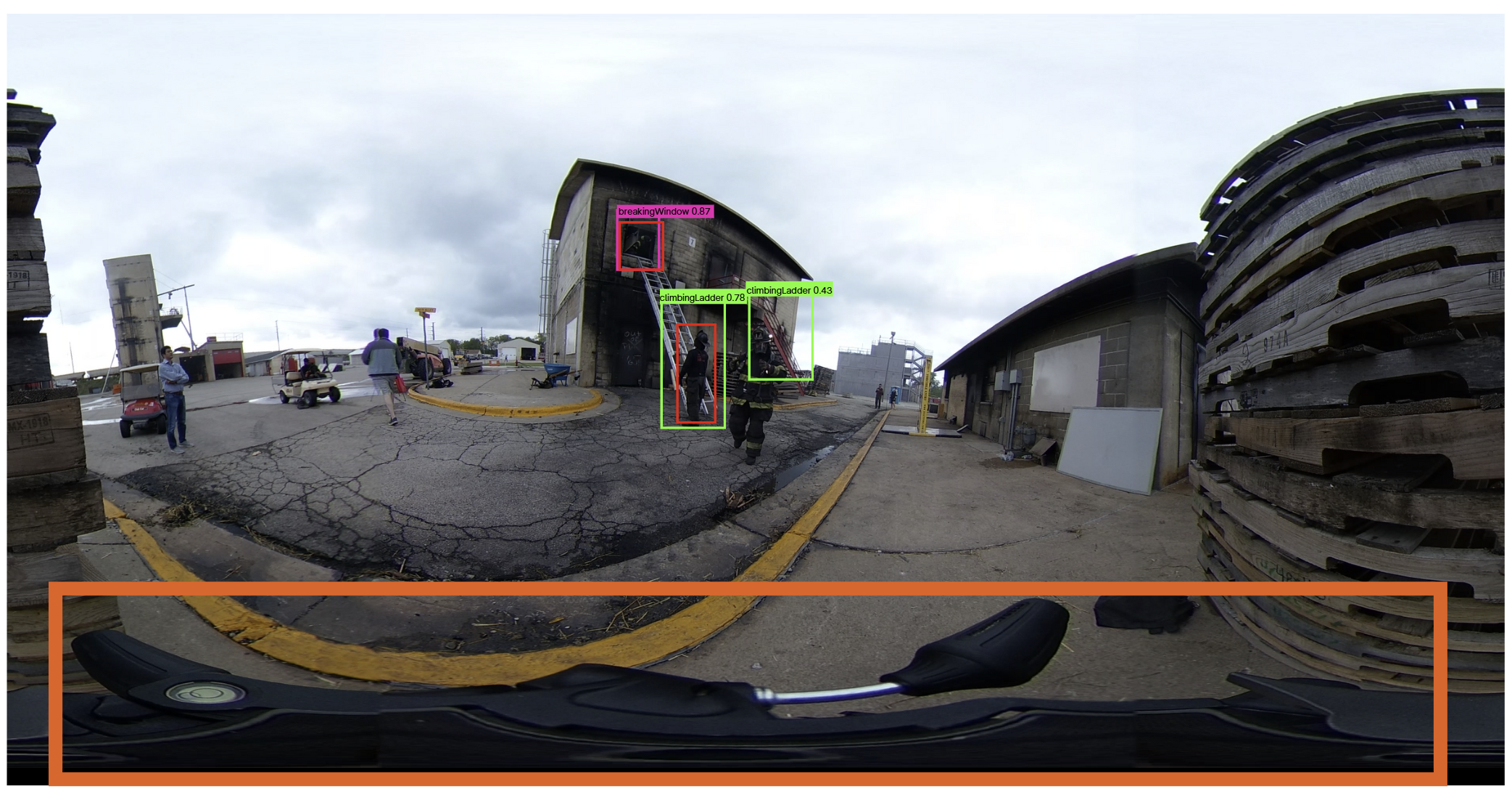}
    \caption{360-degree video frame showing ERP distortion near poles (orange box), demonstrating the need for ERP-aware processing.}
    \label{fig:distortion_ERP}
\end{figure}

\subsection{Related Work} 
Unlike rectilinear (2D) videos, 360-degree video frames suffer from severe ERP distortions, particularly near the poles, as illustrated in Figure~\ref{fig:distortion_ERP}, making action detection more challenging in panoramic settings. Action detection in 2D video has been extensively studied, with deep learning-based models achieving state-of-the-art results. Several approaches have been explored for spatiotemporal action recognition, including SlowFast Networks \cite{slowfast}, which separate spatial and temporal feature extraction, Faster R-CNN \cite{fasterRcnn}, which applies object detection-based techniques to locate actors, and 2D Conv-RBM + LSTM \cite{convRBM}, which integrates spatial and temporal modeling. Among one-stage detectors, YOWO efficiently integrates spatial and temporal features into a single-shot architecture, making it suitable for real-time applications \cite{rwPara1ref5Kpkl2019YouOW}. Several recent models, including InternVideo2 \cite{internvideo} and TokenLearner \cite{tokenlearner}, focus on video understanding but require significant adaptation for panoramic distortions. Similarly, ActionCLIP \cite{actionclip} and EffHandNet \cite{effhandnet} excel in classification and egocentric interactions, respectively, but lack explicit action localization capabilities. While these methods demonstrate strong spatiotemporal modeling, they lack the efficiency required for mission-critical debriefing. The proposed 360YOWO extends YOWO by integrating ERP-aware processing through EAC and 360-degree spatial attention, ensuring robust detection in panoramic settings.

\boldheader{360-Degree Video Processing.} Existing research on 360 degree video understanding focuses mainly object detection and viewport prediction \cite{jiaxi1, jiaxi2} but does not directly address action detection. Spherical CNNs attempt to adapt convolutional layers for ERP distortions\cite{spherecnn}, while multi-view projection methods divide panoramic images into rectilinear views before applying standard CNNs. Viewport prediction models predict user gaze movements in VR applications but lack global action detection capabilities. To handle ERP distortions, prior works have introduced equirectangular convolutions, spherical convolutions \cite{sphericalvisiontransformer}, and mesh-based approaches, which improve object recognition but are not designed for spatiotemporal action detection. 360YOWO integrates EAC, a custom distortion-aware convolutional operation, to ensure accurate feature extraction across panoramic video data.

\boldheader{Action Recognition in 360-Degree Videos.} Research on 360-degree action detection remains limited, with only a few datasets supporting panoramic action recognition. The 360Action dataset \cite{360actiondataset} provides 784 omnidirectional action clips, while CVIP360 and NCTU-GTAV360 datasets \cite{nctu} support tracking and synthetic video analysis. These datasets contribute to 360-degree action recognition research but lack real-world, mission-critical training environments. The firefighting dataset introduced in ACT360 bridges this gap, capturing high-risk training exercises under diverse environmental conditions.

\boldheader{Large Language Models for Summarization.} Transformer-based architectures such as GPT, BERT, and T5 have demonstrated strong capabilities in extracting structured information from videos. However, existing summarization models generate generic captions and lack the structured debriefing necessary for mission-critical applications. ACT360 introduces 360AIE, which leverages GPT-4 \cite{gpt4} fine-tuned on structured firefighter training logs to convert detected actions into scenario-specific debriefing reports, aiding post-training analysis.

\boldheader{Why 360YOWO? Justification for Model Choice.} While alternative architectures like TSN, InternVideo2, and 2D Conv-RBM + LSTM offer advantages in long-term temporal modeling, 360YOWO was selected for its single-stage efficiency, spatiotemporal integration, and fast performance. Unlike TSN and SlowFast, which require multi-stream fusion, YOWO's end-to-end architecture enables efficient detection. InternVideo2 focuses on general video understanding rather than action localization, while 2D Conv-RBM + LSTM, though effective in temporal modeling, lacks YOWO’s efficiency for low-latency spatiotemporal processing. By integrating EAC and spatial attention, 360YOWO extends single-shot action detection to the 360-degree domain, offering a scalable solution for panoramic action recognition.

%% file: sections/3-framework.tex
\begin{figure*}[htbp]
    \centering
    \includegraphics[width=\linewidth, height=0.3\textheight]{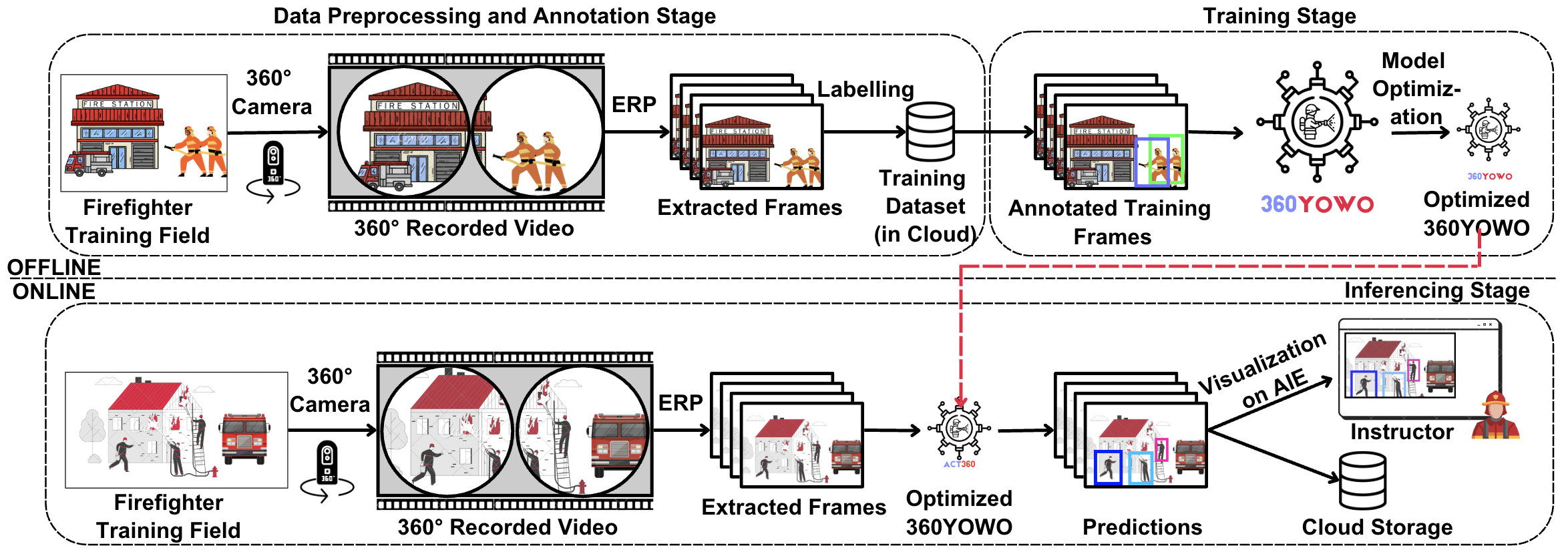}
    \caption{ACT360 framework overview illustrating three main stages: (1) Data Preprocessing, (2) Model Training, and (3) Inference.}
    \label{fig:roadmap}
\end{figure*}

\section{ACT360 Framework}
\label{sec:act360_framework}

\subsection{Overview of ACT360}
ACT360 is an end-to-end framework designed for resource-constrained action detection and debriefing in mission-critical training environments. As illustrated in Figure~\ref{fig:roadmap}, ACT360 consists of three key stages: (1) Data Preprocessing, (2) Model (360YOWO) Training, and (3) Inference. The system integrates 360YOWO for action detection, 360AIE for visualization and summarization, and model optimization for efficient deployment. 360YOWO is trained and evaluated on a firefighter training dataset, which consists of annotated 360-degree videos capturing real-world emergency response scenarios. Further details on the dataset and its composition are provided in Section~\ref{sec:dataset_section}.

\begin{figure*}[t]
\centering
\includegraphics[width=\linewidth]
{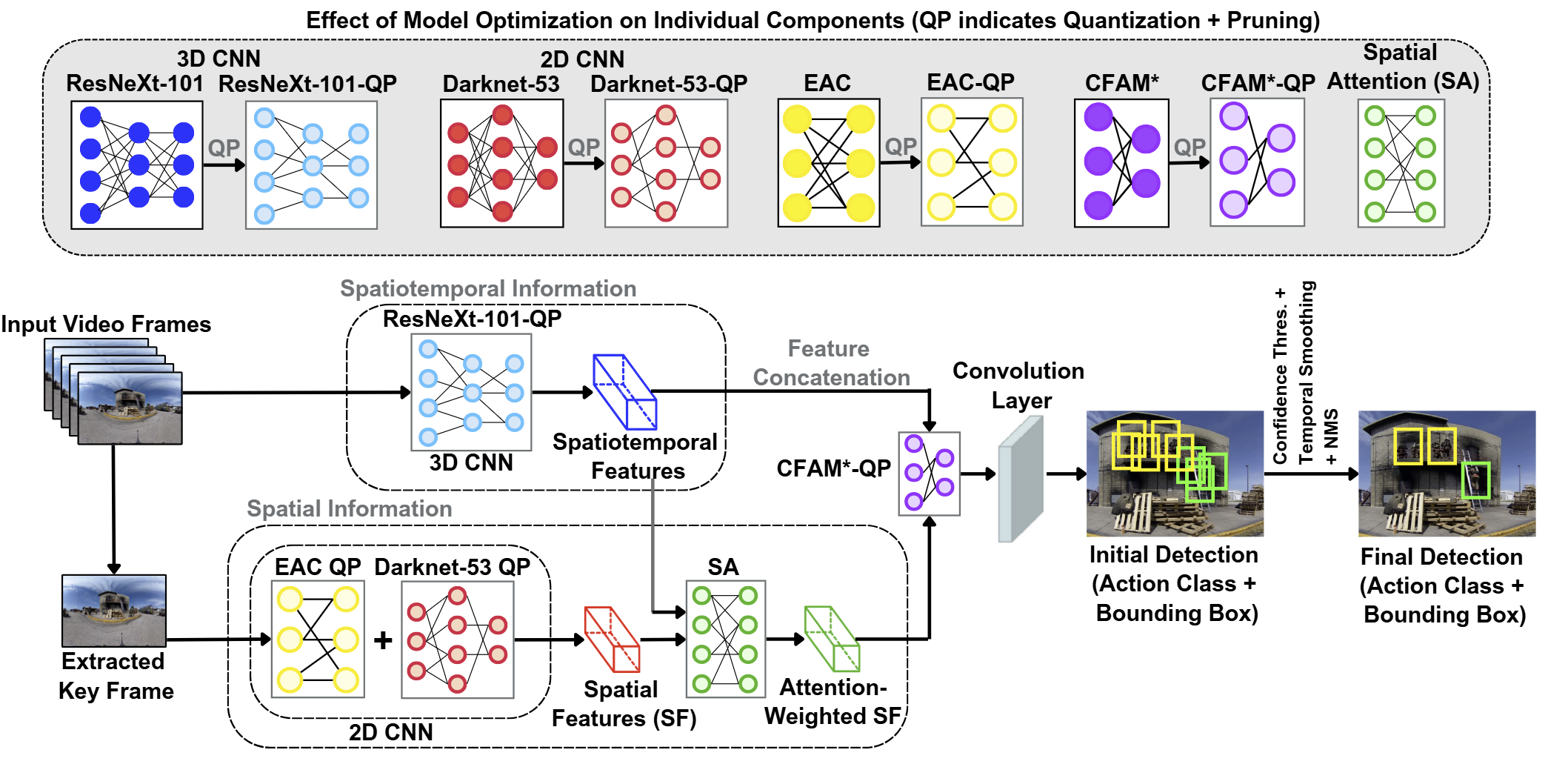}
\caption{360YOWO architecture and optimization pipeline. Top: Effects of quantization (FP32 to INT8) and pruning (QP) on individual components. Bottom: Dual-stream architecture integrating spatiotemporal (3D CNN) and spatial (2D CNN + EAC) processing, refined through CFAM-QP for final action detection.}
\label{fig:ACT360pipeline}
\end{figure*}

\begin{figure}[h]
  \centering
  \includegraphics[width=\linewidth]{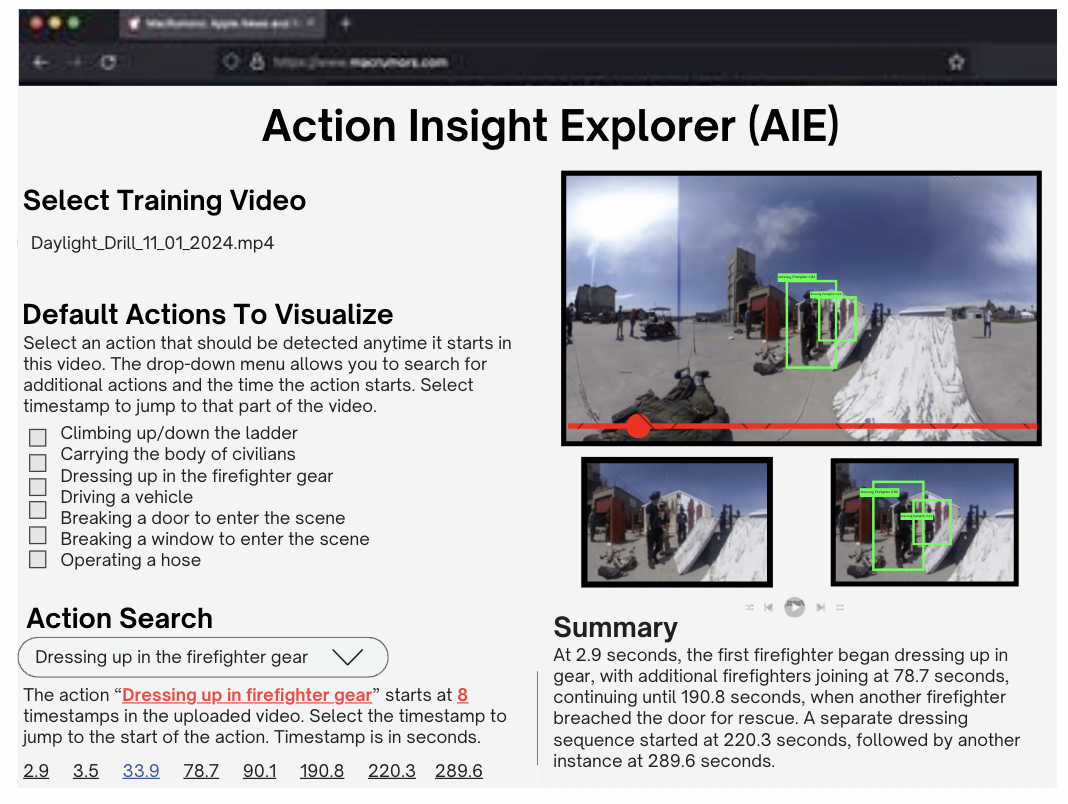}
  \caption{360AIE interface displaying key components, including video selection, detection overlays, zoomed-in action views, an action timeline, and text-based summaries.}
  \label{fig:aieInterface}
\end{figure}

The 360YOWO architecture comprises four key components: (1) the baseline YOWO model for spatiotemporal action detection, (2) Equirectangular-Aware Convolution (EAC) to mitigate ERP distortions, (3) a 360-degree spatial attention mechanism for improved feature weighting, and (4) model optimization techniques (Section~\ref{sec:model_optimization}) to enhance efficiency and deployment. The full architecture is illustrated in Figure~\ref{fig:ACT360pipeline}. 360AIE seamlessly integrates into this pipeline by processing detected actions and generating structured textual summaries, ensuring a smooth transition from raw video input to actionable insights for post-training debriefing. Figure \ref{fig:aieInterface} provides an overview of the 360AIE interface and its key functionalities.

\subsection{Base Model: YOWO Architecture}
\label{subsec:360YOWO}

The foundation of 360YOWO is an enhanced version of \textbf{YOWO} \cite{rwPara1ref5Kpkl2019YouOW}, a single-stage spatiotemporal action detector that integrates 2D CNNs for spatial feature extraction and 3D CNNs for temporal modeling. YOWO is selected for its ability to efficiently process spatial and temporal dependencies, making it well-suited for dynamic firefighter training videos. Its speed and simplicity make it effective for complex action detection and analysis.

However, YOWO was originally designed for 2D rectilinear video and does not account for geometric distortions in ERP images. To adapt YOWO for 360-degree action detection, we introduce EAC and a 360-degree spatial attention mechanism, enabling the model to focus on relevant action regions and compensate for ERP distortions. These two enhancements are essential for ensuring robust detection in a 360-degree view.

\subsection{Enhancements to YOWO for 360-Degree Video}
\label{subsec:enhancements}

\subsubsection{Equirectangular-Aware Convolution (EAC)}
\label{subsubsec:eac}

ERP introduces spatial distortions, particularly near the poles, where standard convolutions struggle due to non-uniform pixel distributions. We propose EAC, a modified convolution operation that dynamically scales kernel weights based on latitude, ensuring consistent feature extraction across the entire panoramic view.

\textbf{Mathematical Formulation:}
\begin{enumerate}
    \item \textbf{Latitude Mapping:}
    \begin{equation}
    \label{eq:latitude_mapping}
    \phi(y) = \pi \left(\frac{y}{H} - \frac{1}{2}\right), \quad \phi(y) \in \left[-\frac{\pi}{2}, \frac{\pi}{2}\right]
    \end{equation}
     where \( H \) is the image height and \( \phi(y) \) represents the latitude of a given pixel.

    \item \textbf{Kernel Adjustment:}
    \begin{equation}
    \label{eq:kernel_adjustment}
    W'(\phi) = W \cdot \cos(\phi)
    \end{equation}
     Convolutional weights \( W \) are adjusted based on latitude to compensate for varying pixel densities.

    \item \textbf{Feature Extraction:}
    \begin{equation}
    \label{eq:feature_extraction}
    Y_{c,y,x} = \sum_{i=1}^{C_{\text{in}}} \sum_{m,n} W'_{c,i,m,n}(\phi(y)) \cdot X_{i,y+m,x+n}
    \end{equation}
     This ensures uniform feature extraction across latitudes, improving action recognition.
\end{enumerate}

\subsubsection{360-Degree Spatial Attention}
\label{subsubsec:SpatialAttention}

Standard attention mechanisms treat all spatial regions equally, ignoring ERP distortions. The 360-degree spatial attention mechanism dynamically weights feature maps based on distortions and motion cues, ensuring the model focuses on the most relevant regions.

\textbf{Attention Computation:}
\begin{equation}
\label{eq:attention_computation}
A_{\text{final}} = \sigma(W_d[F_t \parallel F_{t-1}]) \cdot A_{\text{ERP}}
\end{equation}
where \( A_{\text{ERP}}(y) = \frac{1}{\cos(\phi(y))} \) is the distortion-aware attention map. It also factors in motion queues, which makes the model robust.

\subsection{Action Insight Explorer (360AIE)}
\label{subsec:360AIE}

360AIE is an interactive platform for \textit{visualizing detected actions and generating structured debriefing summaries}. Instructors can query detected actions and receive LLM-generated textual summaries, reducing the need for manual video review, which is costly and labor-intensive.

\subsubsection{Summarization Process Using GPT-4}
360AIE integrates GPT-4 \cite{gpt4} to generate structured summaries from detected action sequences. The summarization pipeline follows these steps:

\begin{enumerate}
    \item \textbf{Input Processing:} The detected actions, timestamps, and confidence scores are extracted.
    \item \textbf{Contextual Sequence Analysis:} GPT-4 processes detected actions in sequence to maintain logical flow.
    \item \textbf{Query-Driven Summarization:} Users can specify a time range or filter actions for a customized summary.
    \item \textbf{Text Generation:} GPT-4 generates structured summaries of key actions, ensuring accuracy and coherence in real-world training contexts.
\end{enumerate}

\textbf{Example Output:}
\begin{quote}
"At 10 seconds, the firefighter climbed the ladder. By 22 seconds, they breached the door and proceeded to rescue a civilian at 30 seconds."
\end{quote}

GPT-4 was fine-tuned using supervised training on firefighter training transcripts and structured debriefing reports. The model was trained on action detection sequences paired with expert-written summaries to enhance domain relevance. 

\section{Model Optimization}
\label{sec:model_optimization}

Given the computational challenges of processing 360-degree video, we optimize 360YOWO to reduce model size and improve efficiency while maintaining detection accuracy. This is achieved through post-training quantization (PTQ) and magnitude-based pruning (MBP), enabling deployment on resource-constrained hardware while preserving detection performance.

\boldheader{Quantization.} We apply PTQ \cite{ptq}, converting model weights and activations from FP32 to INT8 precision. Per-channel quantization is used for convolution layers to retain accuracy, while activation values are globally quantized. A calibration set of 1000 frames determines optimal scaling factors. Our results show minimal accuracy degradation (mAP drop: 0.865 $\rightarrow$ 0.855) while achieving a 72\% reduction in model size (287MB $\rightarrow$ 80MB) and a 27\% improvement in inference time.

\boldheader{Magnitude-Based Pruning.} To further optimize the model, we implement iterative MBP, gradually removing 2\% of the lowest-magnitude weights per iteration. This process is followed by fine-tuning over 10 epochs with a reduced learning rate ($1 \times 10^{-5}$) to recover accuracy. The pruning schedule is designed to preserve essential feature representations while reducing computational redundancy. Pruning reduces the model size to 75MB and improves inference speed by 31.25\% (48ms $\rightarrow$ 33ms per frame), with only a 1.5\% drop in mAP.

\boldheader{Optimization Trade-offs.} While structured pruning (channel pruning) and dynamic quantization were considered, these alternatives resulted in higher accuracy loss (>2.5\% mAP) or increased latency due to on-the-fly quantization overhead. Knowledge distillation yielded similar compression but required extensive retraining, making it impractical for rapid deployment. The combination of PTQ and MBP was selected as the most stable and efficient optimization strategy for 360-degree action detection.

\boldheader{Hardware Considerations.} The optimized 360YOWO model is designed for deployment on GPUs with native INT8 acceleration, such as NVIDIA Tensor Cores (available on T4, A100) and Intel DL Boost. However, on GPUs that lack dedicated INT8 support, such as the RTX 3090, inference falls back to FP16 execution. While FP16 provides efficiency improvements over FP32, it does not match the full performance benefits of INT8 acceleration. A batch size of 1 is maintained for inference to minimize latency, with fallback support for non-INT8 hardware, albeit at reduced efficiency.

\boldheader{Implementation.} The optimization pipeline is implemented using PyTorch’s quantization and pruning utilities. Custom calibration routines ensure robustness to 360-degree distortions. The entire optimization process is applied post-training, requiring only the validation dataset for fine-tuning, making it efficient and reproducible.

\begin{table}[t]
\caption{Optimization strategy for different components of 360YOWO.}
\label{tab:component_optimization}
\centering
\small
\setlength{\tabcolsep}{4pt}  
\begin{tabular}{lccc}
\toprule
\textbf{Component} & \textbf{Quant.} & \textbf{Pruning} & \textbf{Rationale} \\
\midrule
\multicolumn{4}{l}{\textit{2D CNN Branch:}} \\
Darknet-53 & \checkmark & \checkmark & High redundancy \\
EAC & \checkmark & - & Preserve correction \\
\midrule
\multicolumn{4}{l}{\textit{3D CNN Branch:}} \\
ResNeXt-101 & \checkmark & \checkmark & High redundancy \\
\midrule
\multicolumn{4}{l}{\textit{Attention:}} \\
Spatial & FP16 & - & Keep precision \\
Channel & FP16 & - & Keep precision \\
Feature Fusion & \checkmark & \checkmark & Compressible \\
\midrule
\multicolumn{4}{l}{\textit{Detection Head:}} \\
Classification & \checkmark & \checkmark & Acceptable loss \\
Box Regression & \checkmark & \checkmark & Acceptable loss \\
\bottomrule
\end{tabular}
\end{table}

\boldheader{Final Optimized Model.} With quantization and pruning, 360YOWO achieves a 74\% reduction in model size and a 31.25\% speedup in inference time, making it deployable in mission-critical training environments. A detailed evaluation of optimization impact is presented in Section~\ref{sec:exp_results}.

%% file: sections/4-dataset.tex
\section{Dataset}
\label{sec:dataset_section}

\subsection{Motivation and Purpose}
Existing action detection datasets primarily focus on 2D video recordings, making them unsuitable for evaluating models in 360-degree, mission-critical environments. While datasets such as UCF101-24 \cite{UCF101}, and AVA \cite{AVA} provide valuable benchmarks for general action recognition, they lack the panoramic perspective, high-risk scenarios, and dynamic spatial challenges present in mission-critical applications. Although prior datasets such as 360Action \cite{360actiondataset} and NCTU-GTAV360 \cite{nctu} support 360-degree action recognition, they focus on general activities, tracking, or synthetic simulations, making them inadequate for real-world emergency response training.


To address this gap, we introduce a publicly available 360-degree firefighter training dataset designed to evaluate 360YOWO’s performance in real-world mission-critical scenarios. This dataset serves as a validation benchmark for panoramic action detection and enables further research in high-stakes operational training. 


\subsection{Data Collection}

The dataset was collected in collaboration with experienced firefighters and instructors at a first responder training facility. Special permissions were secured for access to the training grounds, allowing for an accurate recording of firefighter drills. Training exercises were recorded using a \textit{Ricoh Theta Z1 51GB 360 Camera}, mounted on tripods at a height of 60–70 inches (152–178 cm) and positioned at a shooting distance of 1–5 meters from the training area. 


The dataset includes a variety of training scenarios that simulate authentic firefighting challenges, including confined space operations, ladder rescues, door and window breaching, victim extrication, and hose operation. Videos were recorded under multiple environmental conditions, including daylight, nighttime, and smoke-filled environments. To ensure diversity, the dataset includes participants of different experience levels and demographics. Data was captured from multiple angles, under varying lighting conditions, and recorded multiple times to reduce bias.

\subsection{Annotation Pipeline}

Each video was annotated in the equirectangular projection (ERP) format, ensuring compatibility with computer vision models designed for panoramic video. The resolution of the ERP frames is 3840 × 1920 pixels (4K), maintaining a 2:1 aspect ratio. The annotations include bounding boxes for localized actions, action labels for classification, and timestamps for temporal detection.

To ensure annotation accuracy, we employed a semi-automated approach using MATLAB's built-in ROI tracking algorithm. The process (Algorithm \ref{alg:annotation_pipeline}) begins with manual region selection, where the first and last frames containing the action are labeled. The MATLAB ROI algorithm is then used to automatically track and propagate these ROIs across intermediate frames, ensuring temporal consistency. This allows for more efficient annotation while minimizing human error. The generated annotations, including spatial coordinates, action labels, and timestamps, are exported in CSV format for further processing. To validate dataset quality, multiple experts reviewed annotations, and inter-annotator agreement was measured using Cohen’s Kappa, achieving a high consistency score. 

\begin{algorithm}[t]
\caption{360-Degree Video Annotation Pipeline}
\label{alg:annotation_pipeline}
\begin{algorithmic}[1]
\State \textbf{Input:} ERP frames $F = \{f_1, \ldots, f_n\}$, Frame rate $fps$
\State \textbf{Output:} Dataset $D$ with action annotations

\State // Manual ROI Selection
\State $f_{start} \gets \text{First frame with action}$
\State $f_{end} \gets \text{Last frame with action}$
\State $roi \gets \text{Manually select action region in } (f_{start}, f_{end})$

\For{each frame $f \in F$}
    \State $bbox \gets \text{TrackROI}(f, roi)$ \Comment{Track selected pixels}
    \State $t \gets \text{Timestamp}(f)$
    \State $D \gets D \cup \{bbox, \text{action}, t\}$
\EndFor

\State Export $D$ to CSV
\State Validate $D$
\Return $D$
\end{algorithmic}
\end{algorithm}

\subsection{Dataset Statistics}

The dataset consists of 55 labeled videos covering key firefighter training exercises. Table~\ref{tab:firefighting_dataset} provides a breakdown of the number of videos per action. The dataset is split into training (70\%), validation (15\%), and test (15\%) sets, ensuring that no videos are shared between splits.

\begin{table}[htbp]
\centering
\caption{Dataset composition: 55 labeled videos across firefighter training actions.}
\label{tab:firefighting_dataset}
\begin{tabular}{lc}
\toprule
\textbf{Action} & \textbf{Number of Videos} \\
\midrule
Climbing up/down the ladder & 14 \\
Carrying a civilian & 12 \\
Dressing in firefighting gear & 15 \\
Driving a vehicle & 19 \\
Breaking a door & 25 \\
Breaking a window & 16 \\
Operating a hose & 37 \\
\bottomrule
\end{tabular}
\end{table}

%% file: sections/5-experimentresults.tex
\section{Experiments and Results}
\label{sec:exp_results}

We evaluate 360YOWO, the action detection model within ACT360, on the firefighter training dataset introduced in Section~\ref{sec:dataset_section}. The dataset consists of 55 labeled 360-degree videos capturing various firefighting activities. It is divided into training (70\%), validation (15\%), and test (15\%) sets, ensuring no video overlap between splits.

Experiments are conducted using both local and cloud-based infrastructures. For training, we use a Dell XPS 15 9530 with an NVIDIA RTX 3090 GPU, where computations are performed in FP16 due to the lack of dedicated INT8 acceleration. Multi-user evaluations are performed on Google Cloud Platform (GCP) using n1-standard-8 instances with NVIDIA T4 GPUs, which support INT8 acceleration for optimized inference.

For inference, batch size is set to 1 to minimize latency. While INT8 quantization provides significant performance gains on supported hardware, the RTX 3090 defaults to FP16 inference, leading to differences in computational efficiency compared to INT8-accelerated GPUs.

The models are trained for 50 epochs using the Adam optimizer with a learning rate of $5 \times 10^{-4}$ and batch size 8. Early stopping is applied based on validation loss.

\boldheader{Evaluation Metrics.} The performance of 360YOWO is measured using three key metrics: Mean Average Precision (mAP) at the video level (consistent with YOWO’s evaluation protocol), Inference Time (processing time per frame in milliseconds), and Model Size (storage size in megabytes (MB)).

\subsection{Baseline Comparisons}
\label{subsec:baseline_comparisons}


Since dedicated 360-degree action detection models are scarce, we evaluate 360YOWO against well-established 2D action detection models adapted for panoramic video. However, these 2D models suffer from ERP distortions, lack specialized spatial attention, and require extensive retraining, making them less effective for mission-critical scenarios. While 360-degree object detection and viewport prediction models incorporate attention mechanisms, they are designed for viewpoint estimation rather than fine-grained action recognition, making them unsuitable as baselines.

We evaluate Faster R-CNN, SlowFast Networks, and YOWO, three commonly used 2D action detection models. Faster R-CNN serves as a strong benchmark as a two-stage detector with high accuracy, while SlowFast Networks capture multi-speed motion patterns. YOWO’s single-stage architecture provides a baseline for comparison.

\boldheader{Adapting 2D Models for 360-Degree Video.}  
To ensure a fair comparison, all baseline models are fine-tuned on the firefighter dataset using ImageNet pre-trained weights and ResNet-50. Adaptations include freezing early convolutional layers while retraining deeper layers to account for 360-degree distortions. Each model follows the same hyperparameters as 360YOWO to maintain consistency.

\begin{figure}[h]
  \centering
  \includegraphics[width=\linewidth]{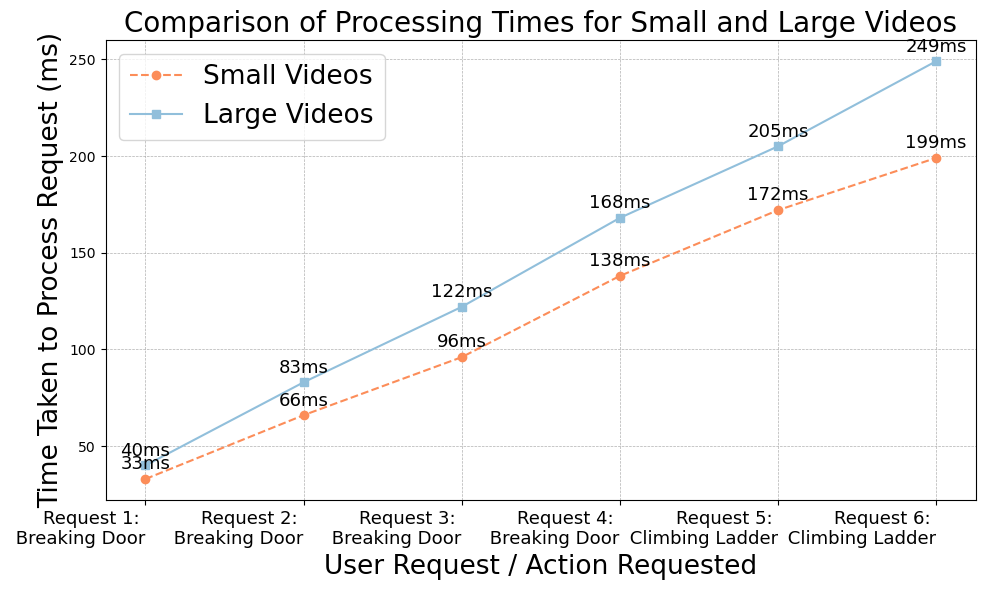}
  \caption{Results of multi-user evaluation where six concurrent requests for different actions were sent to 360AIE. The plot shows the inference times for processing those requests for both short videos (less than six minutes) and long videos (more than ten minutes), highlighting the impact of video length on processing time.}
  \label{fig:comparison}
\end{figure}

\begin{table}
  \caption{Comparison of 360YOWO and baselines in accuracy and inference time.}
  \label{tab:baseline_comparison}
  \centering
  \begin{tabular}{lcc}
    \toprule
    \textbf{Model} & \textbf{mAP} & \textbf{Time (ms)} \\
    \midrule
    YOWO (Adapted) & 0.823 & 46 \\
    Faster R-CNN & 0.781 & 80  \\
    SlowFast & 0.810 & 70 \\ 
    360YOWO & \textbf{0.865} & 48 \\
    360YOWO (Optimized) & 0.850 & \textbf{33} \\
  \bottomrule
\end{tabular}
\end{table}

\begin{table}
  \caption{Ablation Study on the Impact of EAC and Spatial Attention in 360YOWO.}
  \label{tab:ablation_eac_spatial}
  \centering
  \begin{tabular}{lcc}
    \toprule
    \textbf{Configuration} & \textbf{mAP} & \textbf{Time (ms)} \\
    \midrule
    Baseline YOWO (No EAC, No Spatial Attention) & 0.823 & 46 \\
    EAC Applied (No Spatial Attention) & 0.853 & 48 \\
    Full Model (EAC + Spatial Attention) & \textbf{0.865} & 48 \\
    \bottomrule
  \end{tabular}
\end{table}

Table~\ref{tab:baseline_comparison} shows that 360YOWO achieves the highest mAP (0.865) while maintaining an inference speed of 48ms per frame. The optimized version further improves inference time to 33ms, demonstrating the effectiveness of quantization and pruning. Table~\ref{tab:ablation_eac_spatial} presents the ablation study demonstrating the impact of EAC and Spatial Attention on 360YOWO, highlighting significant improvements in mAP while maintaining inference efficiency.

\subsection{Impact of Model Optimization}
\label{subsec:optimization_impact}

\begin{figure}[htbp]
  \centering
  \includegraphics[width=0.8\linewidth, height=0.3\textheight]{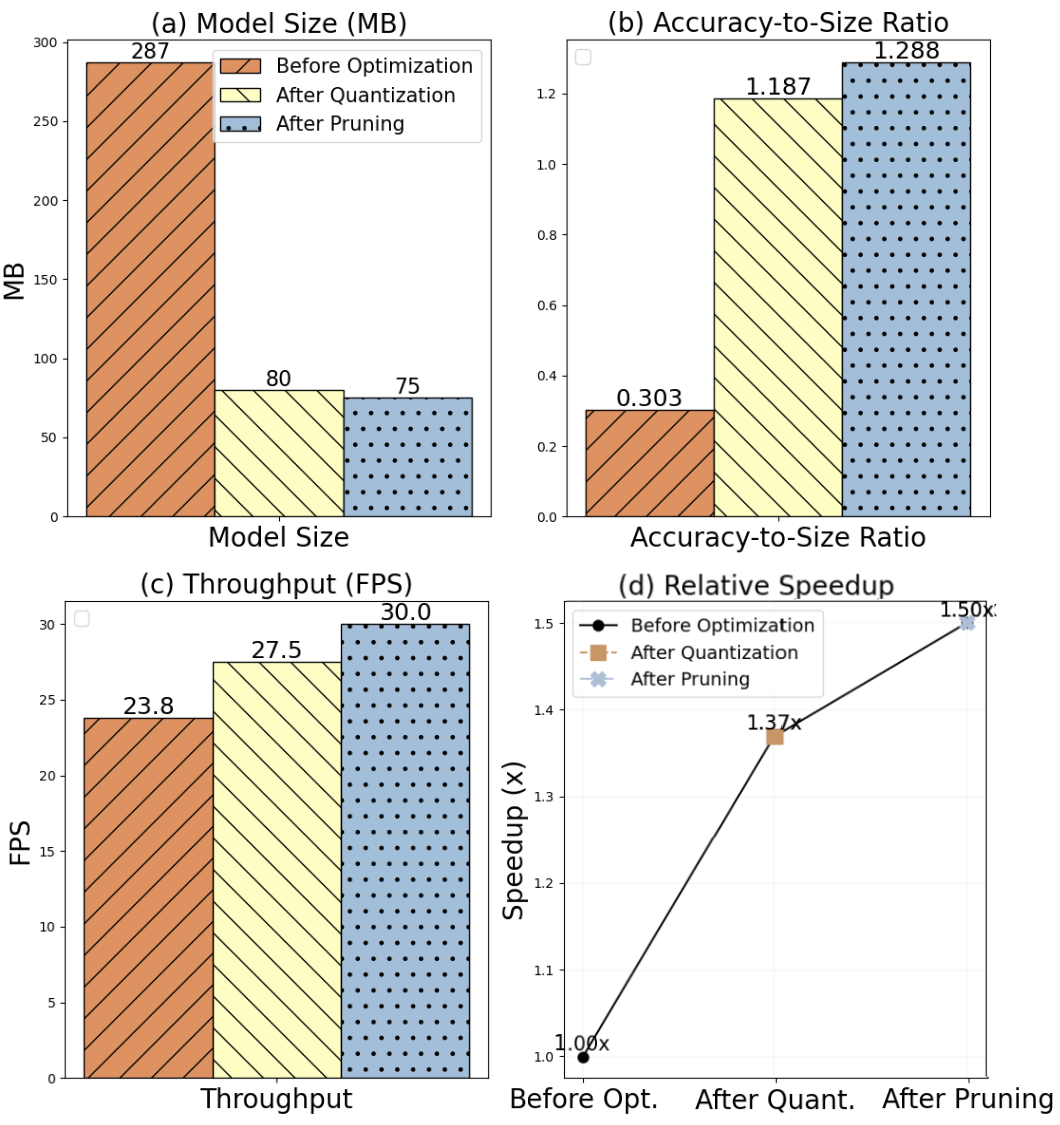}
  \caption{Effect of quantization and pruning on key metrics: (a) Model Size, (b) Accuracy-to-Size Ratio, (c) Throughput (FPS), (d) Speedup.}
  \label{fig:optimization_metrics_results}
\end{figure}

Our optimization pipeline significantly reduces storage and improves inference speed while maintaining detection accuracy. Figure \ref{fig:optimization_metrics_results} shows the effect of model optimization. Quantization reduces model size by 72\% (287MB to 80MB) and inference time by 27\%. Pruning further compresses the model to 75MB, improving inference speed by 31.25\% (48ms to 33ms per frame) with only a 1.5\% drop in mAP.

\begin{table}[ht]
  \caption{Impact of optimizations on mAP, inference time, and model size. Abbreviations: N - Non-Maximum Suppression (NMS), T - Temporal Detection Smoothing, C - Confidence Thresholding, Q - Quantization, and P - Pruning.}
  \label{tab:optimization_impact}
  \centering
  \setlength{\tabcolsep}{4pt} 
  \begin{tabular}{lccc}
    \toprule
    \textbf{Model} & \textbf{mAP} & \textbf{Time (ms)} & \textbf{Size (MB)} \\ 
    \midrule
    360YOWO  & 0.865 &  48 & 287 \\
    + NTC  & 0.865  & 41 & 287  \\
    + NTCQ & 0.855  &  35 & 80  \\
    + NTCQP  & 0.850 & 33 & 75   \\
    \bottomrule
  \end{tabular}
\end{table}

\subsection{Performance Across Conditions}
\label{subsec:qualitative}
\begin{figure}[h]
    \centering
    \includegraphics[width=0.8\linewidth]{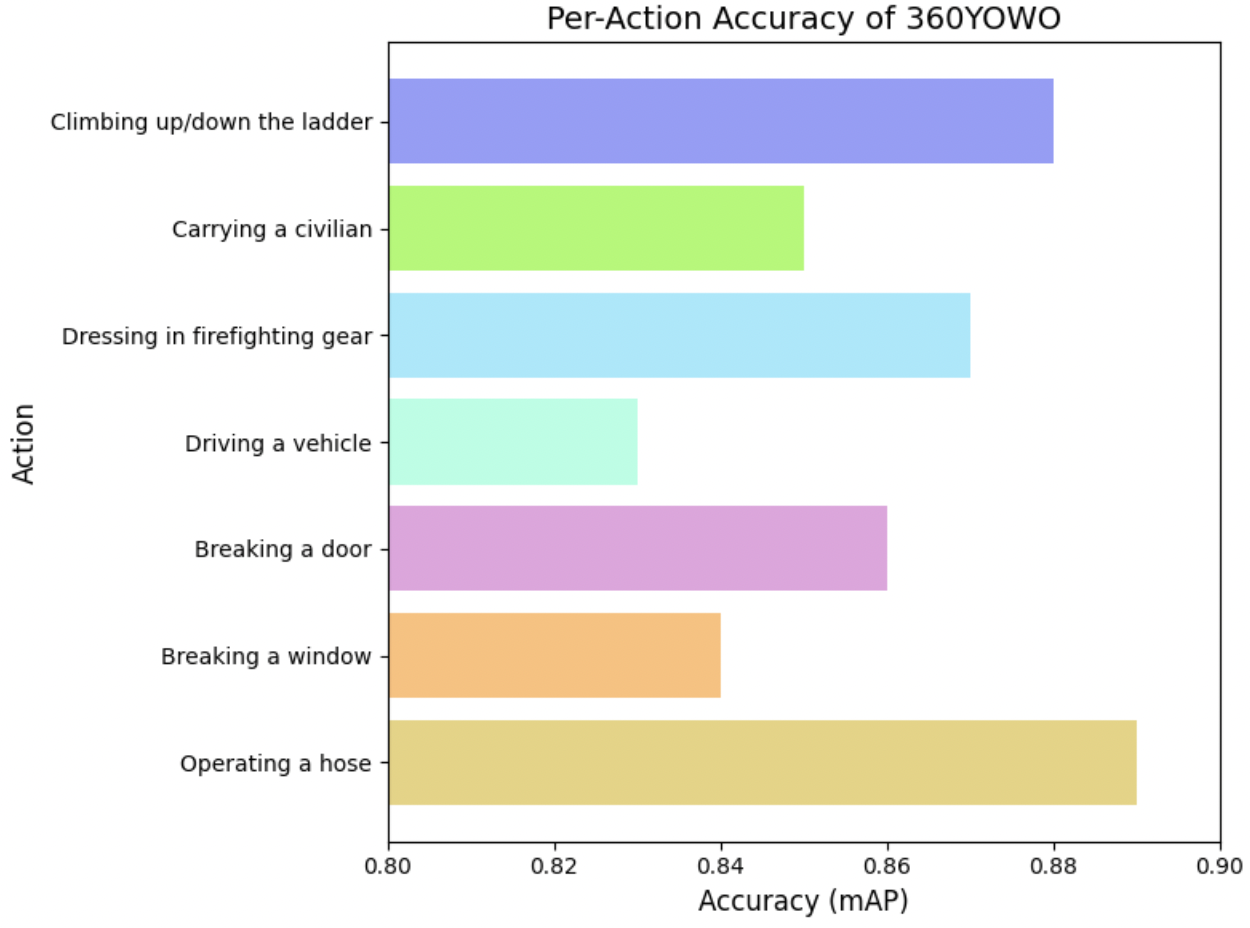}
    \caption{Per-action accuracy (mAP) of 360YOWO across different firefighter activities. The plot illustrates variations in detection performance, highlighting which actions are recognized more accurately by the model.}
    \label{fig:per_action_accuracy}
\end{figure}

\begin{figure}[h]
    \centering
    \includegraphics[width=\linewidth]{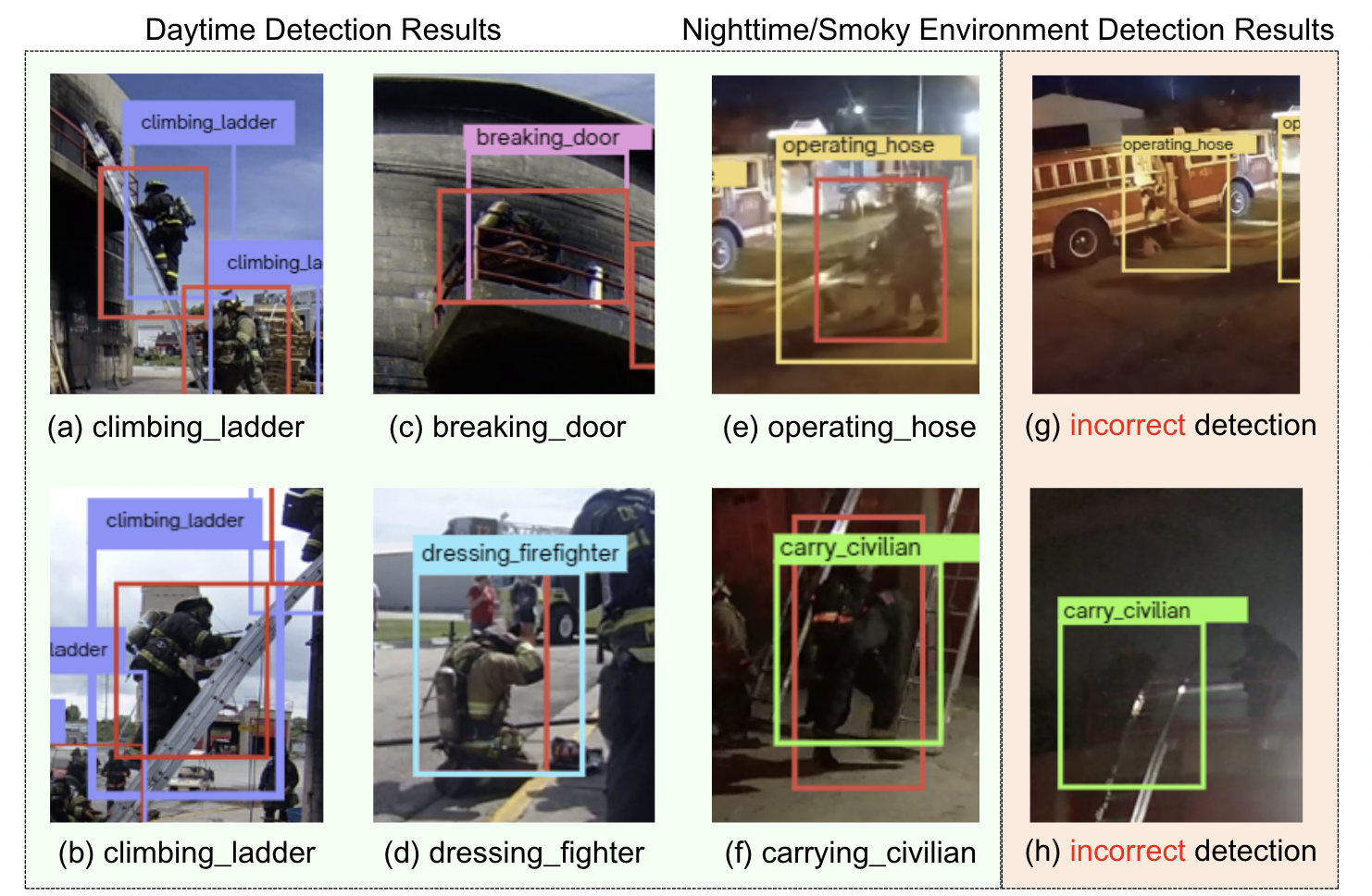}
\caption{360YOWO's performance across conditions. (a-d) Successful detections in daylight, (e-f) correct detections in nighttime/smoky environments, and (g-h) misclassifications in low-visibility conditions due to occlusions and poor lighting. Red boxes denote ground truth; colored boxes indicate predictions.}
    \label{fig:detection_results}
\end{figure}

360YOWO is tested across daylight, nighttime, and smoke-filled environments, as shown in Figure~\ref{fig:detection_results}, achieving an mAP of 0.865 in daylight, 0.781 in nighttime, and 0.629 in smoke-filled conditions. While it performs well in both bright and dark settings, low-visibility conditions lead to misclassifications due to occlusions and reduced contrast. Figure~\ref{fig:per_action_accuracy} presents the per-action accuracy (mAP) scores for 360YOWO, illustrating the model’s performance across different firefighter activities. This breakdown highlights variations in detection accuracy among different actions, providing insight into which activities the model recognizes more accurately. Figure~\ref{fig:comparison} further illustrates the impact of video length on inference time, demonstrating the scalability of 360AIE for handling concurrent action detection requests. These results emphasize the framework’s adaptability for large-scale multi-user debriefing in training environments.

